\documentclass[runningheads]{llncs}
\usepackage[T1]{fontenc}
\usepackage{graphicx}
\usepackage{multirow}
\usepackage[misc]{ifsym}

\usepackage{amssymb}
\usepackage[section]{placeins}
\begin{document}
\title{Enhancing Knowledge Graph Completion with Entity Neighborhood and Relation Context}
\titlerunning{Enhancing KGC with Entity Neighborhood and Relation Context}
%\protect\footnotetext{\textsuperscript{\Letter}Corresponding author.}
\author{Jianfang Chen\inst{1} \and
Kai Zhang\inst{1}\textsuperscript{(\Letter)} \and
Aoran Gan\inst{1} \and
Shiwei Tong\inst{2} \and
Shuanghong Shen\inst{3} \and
Qi Liu\inst{1}}

\authorrunning{J. Chen et al.}
%Tencent Company, Shenzhen, China
\institute{State Key Laboratory of Cognitive Intelligence, University of Science and Technology of China, Hefei, China\\
\email{\{chenjianfang,gar\}@mail.ustc.edu.cn, \{kkzhang08,qiliuql\}@ustc.edu.cn}\and
Tencent Company, Shenzhen, China\\
\email{shiweitong@tencent.com}\and
Institute of Artificial Intelligence, Hefei Comprehensive National Science Center, Hefei, China\\
\email{shshen@iai.ustc.edu.cn}}

\maketitle              % typeset the header of the contribution

\begin{abstract}
Knowledge Graph Completion (KGC) aims to infer missing information in Knowledge Graphs (KGs) to address their inherent incompleteness. Traditional structure-based KGC methods, while effective, face significant computational demands and scalability challenges due to the need for dense embedding learning and scoring all entities in the KG for each prediction. Recent text-based approaches using language models like T5 and BERT have mitigated these issues by converting KG triples into text for reasoning. However, they often fail to fully utilize contextual information, focusing mainly on the neighborhood of the entity and neglecting the context of the relation. To address this issue, we propose KGC-ERC, a framework that integrates both types of context to enrich the input of generative language models and enhance their reasoning capabilities. Additionally, we introduce a sampling strategy to effectively select relevant context within input token constraints, which optimizes the utilization of contextual information and potentially improves model performance. Experiments on the Wikidata5M, Wiki27K, and FB15K-237-N datasets show that KGC-ERC outperforms or matches state-of-the-art baselines in predictive performance and scalability. 

\end{abstract}

\section{Introduction}
Knowledge graphs (KGs) are structured collections of factual data represented as triples $(h, r, t)$, where a head entity $h$ is connected to a tail entity $t$ through a relation $r$. Notable examples include Freebase~\cite{bollacker2008freebase}, ConceptNet~\cite{speer2017conceptnet}, and YAGO~\cite{suchanek2007yago}. These KGs have been extensively utilized in various applications, ranging from recommendation systems and question answering to information retrieval. Their utility has further expanded to support complex AI tasks, including aspect-level sentiment analysis~\cite{zhang2021eatn} and adaptive learning prediction~\cite{liu2019ekt}. However, the inherent incompleteness of KGs often compromises the accuracy of downstream applications, highlighting the critical need for effective Knowledge Graph Completion (KGC) techniques.

Current KGC methods are broadly categorized into structure-based and text-based approaches. Structure-based methods map entities and relations into low-dimensional vectors and evaluate triple plausibility using scoring functions such as TransE~\cite{bordes2013translating}, DistMult~\cite{yang2015embedding}, ComplEx~\cite{trouillon2016complex}, and ROTATE~\cite{sun2018rotate}. These methods capture relational patterns effectively but face scalability challenges due to dense embedding learning and the need to score all entities for each prediction. In contrast, text-based methods like SimKGC~\cite{wang2022simkgc}, KGT5~\cite{saxena2022sequence}, SKG-KGC~\cite{shan2024multi}, and KnowC~\cite{yang2024knowledge} transform KG triples into text and use language models for reasoning, improving scalability. However, existing text-based KGC methods often neglect the full utilization of contextual information, focusing mainly on the \textit{entity neighborhood} while ignoring the \textit{relation context}. These contexts are crucial for language model inference.

To address this limitation, we introduce KGC-ERC, a novel framework that incorporates both entity neighborhood and relation context as ancillary cues within generative language models. When predicting a specific triple, we extract the surrounding entities and triples involving the given relation as context. These contextual information are merged with the target triple and input into the language model, enriching the entities and relations with additional relevant information and improving inference accuracy. However, due to the input token limitations of language models, not all contextual information can be accommodated. To solve this, we propose a sampling strategy for both entity neighborhood and relation context. For entity neighborhood, we collect direct neighbors with diverse relationships to gather a broader range of entity attributes. For relation context, we classify relations into four types (1-n, n-1, n-n, 1-1) based on cardinality and design distinct sampling strategies for each type to ensure the collected context is useful. 

We evaluate KGC-ERC on Wikidata5M~\cite{wang2021kepler}, Wiki27K~\cite{lv2022pre}, and FB15K-237-N~\cite{lv2022pre} datasets, comparing MRR and Hits@$k$ metrics against state-of-the-art methods. Results show that KGC-ERC achieves superior performance. In summary, our contributions include:
\begin{itemize}
  \item We introduce a novel framework that integrates entity neighborhood and relation context into the input of a generative language model, enhancing its inferential capabilities.
  \item We propose a sampling method for efficiently selecting entity neighborhood and extracting relevant relation context, ensuring that our model can process and utilize the most pertinent information within the input capacity constraints.
  \item KGC-ERC is evaluated on three benchmark datasets: Wikidata5M, Wiki27K, and FB15K-237-N. The results show that our model outperforms existing baselines in predictive performance and scalability.
\end{itemize}

\section{Preliminaries}
\label{preliminaries}

\subsubsection{Knowledge Graph Completion}
Knowledge graphs (KGs) are structured representations of facts about entities and their relations, formally defined as a set of triples $\mathcal{G} = \{(h, r, t)\}$, where $h \in \mathcal{E}$ is the head entity, $r \in \mathcal{R}$ is the relation, and $t \in \mathcal{E}$ is the tail entity. Here, $\mathcal{E}$ and $\mathcal{R}$ denote the sets of entities and relations, respectively. Knowledge Graph Completion (KGC) aims to predict missing triples in a KG using existing information. For incomplete triples such as $(h, r, ?)$, the goal is to identify the most probable entity to complete the triple.

\subsubsection{Sequence-to-sequence KGC}
Seq2seq KGC treats the task as a text generation problem using generative language models. Instead of ranking all possible entities, the model directly generates the missing entity as a sequence. For example, the incomplete triple $(StylipS, genre, ?)$ is transformed into a textual query $q$: ``$query: StylipS | genre$''. To enrich the input, context from \textit{Entity Neighborhood} ($E_n$) and \textit{Relation Context} ($R_c$) is appended to form:
\begin{equation}
q' = concat(q, E_n(q), R_c(q))
\end{equation}
This enriched input $q'$ is then passed to the model to generate the output tokens:
\begin{equation}
T_{out} = Decoder(Encoder(q'))
\end{equation}
The goal is to generate the missing entity. Following the methodology of KGT5-context~\cite{kochsiek2023friendly}, we create an inverse triple $(t, r^{-1}, h)$ for each original triple $(h, r, t)$, where $r^{-1}$ denotes the reciprocal relation. This approach focuses on predicting the tail entity while considering bidirectional relations.

\section{KGC-ERC Framework}
\begin{figure}[tb]
\centering
\includegraphics[width=\textwidth, keepaspectratio]{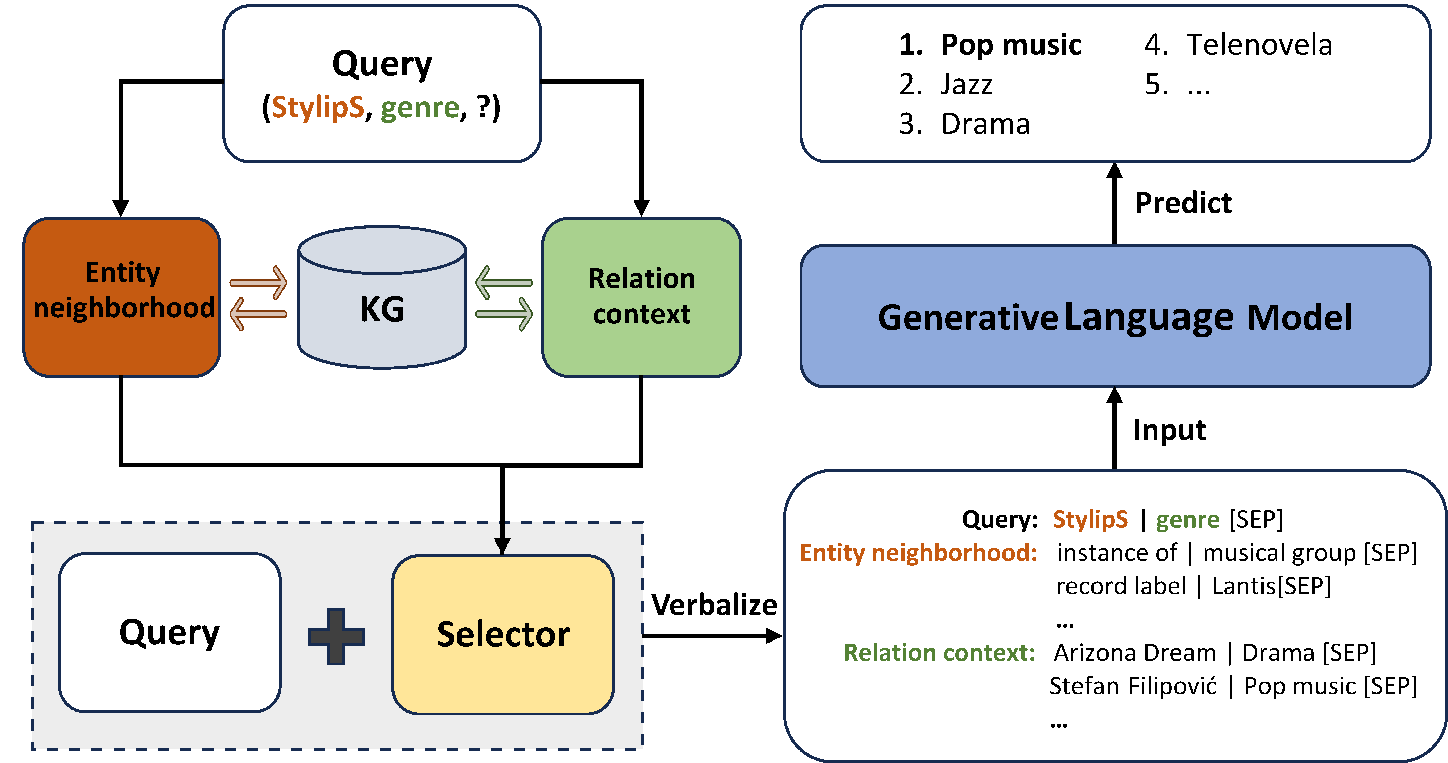}
\caption{Illustration of the KGC-ERC framework for knowledge graph completion. Real example from Wikidata5M dataset, best viewed in color.} 
\label{KGC-ERC}
\end{figure}

KGC-ERC enhances KGT5-context~\cite{kochsiek2023friendly} by adding relation context to the query, enriching the input sequence and expanding contextual information beyond the entity neighborhood. Given input token limits, a \textit{Selector} module samples relevant entity and relation context, filtering out less important data. As illustrated in Fig.~\ref{KGC-ERC}, it retrieves entity and relation context for a triple with a missing entity, selects relevant context, appends it to the query, and feeds the augmented query into the language model for prediction.

\subsection{Selector}
\label{selector}
The \textit{Selector} module is a critical component of the KGC-ERC model, managing the vast amount of contextual information in the knowledge graph. It ensures the generative language model receives only the most relevant context through Entity Neighborhood Sampling and Relation Context Sampling. Inspired by the interactive attention mechanism proposed by Zhang et al.~\cite{zhang2019interactive}, this dual interaction enhances the model's understanding of the query and captures complex relationships within the knowledge graph.

\subsubsection{Entity Neighborhood Sampling.}
Entity neighborhood sampling identifies all direct neighbors of the target entity, forming a set of relation-entity pairs $(r, e)$. These pairs are grouped by relation and sorted by the size of each group in descending order. We then randomly select one pair from each group, up to a maximum of 50 pairs. This strategy ensures diverse attribute coverage and richer semantic representation. For example, for the entity \textit{LeBron James} with neighbors (\textit{child, Bryce Maximus James}), (\textit{child, Bronny James}), and (\textit{occupation, Basketball player}), selecting pairs from both $child$ and $occupation$ relations provides a more comprehensive representation than selecting only from $child$ relations.

\subsubsection{Relation Context Sampling.}
For relation context sampling, we focus on triples containing the same relation as the query. For 1-n relations, we prioritize triples involving both the head entity $h$ and the relation $r$ from the query. For n-1 relations, we ensure diversity in the tail entities to avoid repetition. For n-n relations, we combine the strategies of the previous two cases, selecting half of the triples based on each approach. For 1-1 relations, random sampling is employed. The maximum number of selected triples is capped at 50.

It is important to note that the information derived from the above two sampling processes explicitly excludes the target triple itself, as failure to do so would render the training task insignificant.

\subsection{Verbalization}
\label{verbalization}
To construct the input for the generative language model, we concatenate the query with the entity neighborhood and the relation context selected by the \textit{Selector} (\S~\ref{selector}). The verbalization process transforms the query and contextual information into a coherent text sequence formatted as follows: ``query: <query head entity> | <query relation> <SEP> entity neighborhood: <neighborhood 1 relation> | <neighborhood 1 tail entity> <SEP> ... <SEP> relation context: <context 1 head entity> | <context 1 tail entity> <SEP> ...''\footnote{When entity descriptions are accessible, we incorporate them by adding ``description: <description of query head entity>''}. A real-world example is given in Fig.~\ref{KGC-ERC}. The final combined input is truncated to a maximum of 512 tokens to fit the input requirements of the language model. 

\begin{table}[tb]
\caption{Statistics of the datasets used in this paper.}
\label{dataset3}
\centering
\setlength{\tabcolsep}{6pt} % 调整列之间的间距
% \resizebox{\textwidth}{!}
% {%
\begin{tabular}{lccccc}
\hline
              & Entities    & Relations   & Train        & Valid      & Test   \\
\hline
FB15K-237-N   & 13,104      & 93          & 87,282       & 7,041      & 8,226  \\
Wiki27K       & 27,122      & 62          & 74,793       & 10,121     & 10,122 \\
Wikidata5M    & 4,594,485   & 822         & 20,614,279   & 5,163      & 5,133  \\
\hline
\end{tabular}
% }
\end{table}

\subsection{Training and Inference}
Training involves constructing $(input, output)$ sequences from training triples $(h, r, t)$, where inputs are queries and context in text format, and outputs are missing entities. We use teacher forcing and cross-entropy loss, without explicit negative sampling. Instead, each token is assessed step-by-step against all possibilities, simplifying the process.

For inference, given a query $q$, we verbalize $q$ and its context, then input it into the model. The generative language model generates candidate entities via autoregressive decoding. This process is repeated multiple times to generate a set of candidate entities. Candidate sequences are scored using the method from KGT5~\cite{saxena2022sequence}, maximizing the likelihood of correct generation. We discard outputs that do not map to valid entity mentions and remove any duplicate entities from the candidate set to ensure uniqueness. This ensures that only a subset of potential entities (far fewer than the total number in $\mathcal{E}$) need to be scored and ranked. If the correct entity is missing from the candidate set, it is assigned a score of $-\infty$, allowing the use of conventional KGC evaluation metrics.

\section{Experiments}

\subsection{Experiment Setup}
\subsubsection{Datasets.} 
We evaluate KGC-ERC on FB15K-237-N~\cite{lv2022pre}, Wiki27K~\cite{lv2022pre}, and Wikidata5M~\cite{wang2021kepler} datasets. Detailed statistics are provided in Table~\ref{dataset3}. FB15K-237-N, derived from FB15K-237~\cite{toutanova2015observed}, contains approximately 13k entities and 93 relations. Wiki27K, constructed from Wikidata, includes around 27k entities and 62 relations. In contrast, Wikidata5M, combining Wikidata with Wikipedia pages, encompasses nearly 5 million entities and 20 million triples.

\subsubsection{Evaluation Metrics.}
We evaluate KGC-ERC using the entity ranking task. For each test triple $(h, r, t)$, the model predicts the tail entity $t$ given the head entity $h$ and relation $r$. Head entity prediction is achieved by reversing the triple. We sample from the decoder, discarding invalid entity mentions.

We use four metrics: mean reciprocal rank (MRR) and Hits@$k$ (H@$k$) for $k \in \{1, 3, 10\}$. MRR measures the mean of the reciprocal ranks of the correct entities, while H@$k$ indicates the proportion of correct predictions within the top-$k$ results. In case of ties, we use the mean rank to avoid misleading results~\cite{sun2020re}. Both MRR and H@$k$ are calculated under the filtered setting~\cite{bordes2013translating}, which excludes known true triples from the training, validation, and test sets. We report the average of these metrics for both head and tail entity predictions.

\begin{table}[tb]
\caption{Experiment results on Wikidata5M dataset. Results of $[\dag]$ are from KGT5~\cite{saxena2022sequence} and the other results are from the corresponding papers. The first group omits textual information, the second group employs mention names, the third group additionally entity descriptions. Best per group underlined, best overall bold.}
\label{wd5M}
\centering
% \small
\setlength{\tabcolsep}{4pt} % 调整列之间的间距
% \resizebox{\textwidth}{!}{%
\begin{tabular}{lccccc}
\hline
\textbf{Model} & \textbf{MRR} & \textbf{H@1} & \textbf{H@3} & \textbf{H@10} & \textbf{Params} \\
\hline
\multicolumn{2}{l}{\textbf{Structure-based Methods}}  & & & & \\ 
\hline
TransE~\cite{bordes2013translating}~\dag & 0.253 & 0.170 & 0.311 & 0.392 & 2.4B \\
DistMult~\cite{yang2015embedding}~\dag & 0.253 & 0.209 & 0.278 & 0.334 & 2.4B \\
SimplE~\cite{kazemi2018simple}~\dag & 0.296 & 0.252 & 0.317 & 0.377 & 2.4B \\
RotatE~\cite{sun2018rotate}~\dag & 0.290 & 0.234 & \underline{0.322} & 0.390 & 2.4B \\
QuatE~\cite{zhang2019quaternion}~\dag & 0.276 & 0.227 & 0.301 & 0.359 & 2.4B \\
ComplEx~\cite{trouillon2016complex}~\dag & \underline{0.308} & \underline{0.255} & - & \underline{0.398} & \underline{614M} \\ 
\hline
\multicolumn{2}{l}{\textbf{Text-based Methods}} & & & &  \\ 
\hline
KEPLER~\cite{wang2021kepler} & 0.210 & 0.173 & 0.224 & 0.277 & 125M \\
MLMLM~\cite{clouatre2021mlmlm} & 0.223 & 0.201 & 0.232 & 0.264 & 355M \\
KGT5~\cite{saxena2022sequence} & 0.300 & 0.267 & 0.318 & 0.365 & \underline{\textbf{60M}} \\
ReSKGC~\cite{yu2023retrieval} & 0.363 & 0.334 & 0.386 & 0.416 & \underline{\textbf{60M}} \\
KGT5-context~\cite{kochsiek2023friendly} & 0.378 & 0.350 & 0.396 & 0.427 & \underline{\textbf{60M}} \\
KGC-ERC (ours) & \underline{0.386} & \underline{0.360} & \underline{0.403} & \underline{0.436} & \underline{\textbf{60M}} \\
\hline
\multicolumn{2}{l}{\textbf{Text-based Methods With Descriptions}} & & & &  \\ 
\hline
SimKGC $+$ Desc.~\cite{wang2022simkgc} & 0.358 & 0.313 & 0.376 & 0.441 & 220M \\
SKG-KGC $+$ Desc.~\cite{shan2024multi} & 0.366 & 0.323 & 0.382 & 0.446 & 220M \\
KGT5-context $+$ Desc.~\cite{kochsiek2023friendly} & 0.426 & 0.406 & 0.440 & 0.460 & \underline{\textbf{60M}} \\
KnowC $+$ Desc.~\cite{yang2024knowledge} & 0.426 & 0.373 & 0.447 & \underline{\textbf{0.525}} & 220M \\
KGC-ERC $+$ Desc. (ours) & \underline{\textbf{0.433}} & \underline{\textbf{0.412}} & \underline{\textbf{0.448}} & 0.467 & \underline{\textbf{60M}} \\
\hline
\end{tabular}%
% }
\end{table}

\subsubsection{Implementation Details.}
We follow the settings of KGT5-context~\cite{kochsiek2023friendly} and use T5~\cite{raffel2020exploring} models for our experiments. For the large-scale Wikidata5M dataset, we train the T5-small model from scratch due to the substantial amount of data. For the smaller FB15K-237-N and Wiki27K datasets, we fine-tune the T5-base model. All models use the SentencePiece tokenizer~\cite{raffel2020exploring}. Details are as follows:

\begin{itemize}
    \item \textbf{Wikidata5M}: Trained on 4 Nvidia 4090 GPUs with a batch size of 64 (effective batch size 256). The optimizer is AdaFactor, and training lasts for 6 epochs. Contextual information (entity neighborhood and relation context) is sampled up to 50 instances each.
    \item \textbf{FB15K-237-N and Wiki27K}: Trained on a single Nvidia A100 GPU with a batch size of 16. The optimizer is AdamW with a learning rate of $5 \times 10^{-5}$ after testing multiple values. Training lasts for 10 epochs. Contextual information is sampled up to 10 entity neighbors and 50 relation contexts.
\end{itemize}

The sampling sizes for entity neighbors and relation contexts were determined through extensive experiments to optimize model performance, ensuring that the chosen numbers are data-driven rather than arbitrary. For inference, we sample 500 entities from the decoder across all datasets.

\subsection{Experiment Results}
\subsubsection{Baselines.}
We compare our KGC-ERC model with various baselines across the three datasets. For Wikidata5M, we include structure-based methods such as TransE, DistMult, SimplE, RotatE, QuatE, and ComplEx, as well as text-based methods like KEPLER, MLMLM, KGT5, ReSKGC, KGT5-context, SimKGC, SKG-KGC, and KnowC. Notably, KGT5, ReSKGC, KGT5-context, and KGC-ERC utilize generative language models. For Wiki27K and FB15K-237-N, we compare with structure-based methods (TransE, ConvE, TuckER, RotatE) and text-based models (KG-BERT, LP-RP-RR, PKGC, VKGC), where PKGC and VKGC incorporate entity descriptions.

\subsubsection{Main Results.}
As shown in Table~\ref{wd5M}, our KGC-ERC method achieves superior performance across the Wikidata5M, Wiki27K, and FB15K-237-N datasets, outperforming existing baselines in nearly all evaluation metrics. On Wikidata5M, KGC-ERC achieves an MRR of 0.386 and H@1 of 0.360, outperforming KGT5-context and ReSKGC. When incorporating entity descriptions, KGC-ERC further improves to an MRR of 0.433 and H@1 of 0.412, outperforming the best baseline, KnowC, in all but H@10. Importantly, KGC-ERC achieves these results with only 60M parameters, demonstrating its efficiency compared to traditional structure-based methods that often require billions of parameters.

In Table~\ref{WFdata}, which reports results on Wiki27K and FB15K-237-N, KGC-ERC again surpasses both structure-based and text-based methods, achieving the highest MRR, H@1 and H@3. Specifically, our method achieves an MRR of 0.305 and 0.343 on Wiki27K and FB15K-237-N, respectively. This shows its robust performance on different benchmarks.

\begin{table}[htb]
\caption{Experiment results on Wiki27K and FB15K-237-N datasets. Results of $[\dag]$ are from PKGC~\cite{lv2022pre} and the other results are from the corresponding papers. The best score is in bold.}
\label{WFdata}
\centering
\setlength{\tabcolsep}{1pt} % 调整列之间的间距
% \resizebox{\textwidth}{!}{%
\begin{tabular}{lcccccccc}
\hline
\multirow{2}*{\textbf{Model}} & \multicolumn{4}{c}{\textbf{Wiki27K}} & \multicolumn{4}{c}{\textbf{FB15K-237-N}} \\
~& \textbf{MRR} & \textbf{H@1} & \textbf{H@3} & \textbf{H@10} & \textbf{MRR} & \textbf{H@1} & \textbf{H@3} & \textbf{H@10} \\ 
\hline
\textbf{Structure-based Methods} & & & & & & & & \\ 
\hline
TransE~\cite{bordes2013translating}~\dag & 0.155 & 0.032 & 0.228 & 0.378 & 0.255 & 0.152 & 0.301 & 0.459 \\
ConvE~\cite{dettmers2018convolutional}~\dag & 0.226 & 0.164 & 0.244 & 0.354 & 0.273 & 0.192 & 0.305 & 0.429 \\
TuckER~\cite{balavzevic2019tucker}~\dag & 0.249 & 0.185 & 0.269 & 0.385 & 0.309 & 0.227 & 0.340 & 0.474 \\
RotatE~\cite{sun2018rotate}~\dag & 0.216 & 0.123 & 0.256 & 0.394 & 0.279 & 0.177 & 0.320 & 0.481 \\\hline
\textbf{Text-based Methods} & & & & & & & & \\ 
\hline
KG-BERT~\cite{yao2019kg}~\dag & 0.192 & 0.119 & 0.219 & 0.352 & 0.203 & 0.139 & 0.201 & 0.403 \\
LP-RP-RR~\cite{kim2020multi}~\dag & 0.217 & 0.138 & 0.235 & 0.379 & 0.248 & 0.155 & 0.256 & 0.436 \\
PKGC~\cite{lv2022pre} & 0.285 & 0.230 & 0.305 & 0.409 & 0.332 & 0.261 & 0.346 & 0.487 \\
VKGC~\cite{han2023knowledge} & 0.296 & 0.241 & 0.320 & \textbf{0.422} & 0.338 & 0.266 & 0.369 & \textbf{0.489} \\
KGC-ERC (ours) & \textbf{0.305} & \textbf{0.267} & \textbf{0.322} & 0.381 & \textbf{0.343} & \textbf{0.280} & \textbf{0.375} & 0.465 \\
\hline
\end{tabular}
% }
\end{table}

\subsection{Ablation Study}
To evaluate the impact of key components in the KGC-ERC model, we conducted ablation studies focusing on the relation context and the customized sampling strategy. Specifically, we tested the model without relation context (using only entity mentions, relation mentions, and 1-hop neighborhood information) and with a random sampling strategy instead of the customized one. The results (Table~\ref{Ablation}) show that excluding the relation context reduces the MRR from 0.386 to 0.375 and H@1 from 0.360 to 0.348. Similarly, replacing the customized sampling strategy with random sampling degrades performance, with the MRR falling to 0.382 and H@1 to 0.356. These findings highlight the importance of both components in enhancing the model's reasoning abilities and optimizing the use of the limited input window. The integration of relation context and the tailored sampling strategy are crucial for achieving state-of-the-art performance and ensuring robust generalizability.

\begin{table}[tb]
\caption{The ablation results on the Wikidata5M dataset.}
\label{Ablation}
\centering
\setlength{\tabcolsep}{12pt} % 调整列之间的间距
\begin{tabular}{lcccc}
\hline
\textbf{Model} & \textbf{MRR} & \textbf{H@1} & \textbf{H@3} & \textbf{H@10} \\
\hline
w/o relation context & 0.375 & 0.348 & 0.392 & 0.425 \\
w/o sampling strategy & 0.382 & 0.356 & 0.398 & 0.431 \\
KGC-ERC & \textbf{0.386} & \textbf{0.360} & \textbf{0.403} & \textbf{0.436} \\
\hline
\end{tabular}
\end{table}

\section{Conclusion}
We introduced KGC-ERC, a novel framework for knowledge graph completion that integrates entity neighborhood and relation context into a generative language model. Our approach achieved superior performance across multiple datasets, outperforming existing baselines in most metrics. Ablation studies demonstrated the importance of relation context and a tailored sampling strategy in enhancing model effectiveness and optimizing the use of the input window. Future work will focus on developing adaptive sampling techniques to dynamically select context more aligned with the query, thereby improving predictive accuracy and broadening the model's applicability.

\begin{credits}
\subsubsection{\ackname} This research was partially supported by the National Natural Science Foundation of China (No.62406303), Anhui Provincial Natural Science Foundation (2308085QF229), the Fundamental Research Funds for the Central Universities (WK2150110034) and the Iflytek joint research program.
\end{credits}

\bibliographystyle{splncs04}
\bibliography{mybibliography}

\end{document}